\documentclass[conference]{IEEEtran}
\IEEEoverridecommandlockouts
\usepackage{cite}
\usepackage{amsmath,amssymb,amsfonts}
\usepackage{algorithmic}
\usepackage{graphicx}
\usepackage{textcomp}
\usepackage{xcolor}

\usepackage{cuted}
\usepackage{graphicx}
\usepackage{cite}
\usepackage{picinpar}
\usepackage{amsmath}
\usepackage{url}
\usepackage[utf8]{inputenc}
\usepackage{colortbl}
\usepackage{soul}
\usepackage{multirow}
\usepackage{pifont}
\usepackage{color}
\usepackage[dvipsnames]{xcolor}
\usepackage{alltt}
\usepackage{hyperref}
\usepackage{enumerate}
\usepackage{siunitx}
\usepackage{breakurl}
\usepackage{epstopdf}
\usepackage{pbox}

\usepackage[ruled,linesnumbered, lined, noend]{algorithm2e}
\usepackage{booktabs}
\newcommand{\tabincell}[2]{\begin{tabular}{@{}#1@{}}#2\end{tabular}} 
\usepackage{bm}
\usepackage{amssymb}
\usepackage{romannum}
\usepackage{bbm}
\usepackage[caption=false,font=footnotesize]{subfig}

\usepackage{booktabs} 
\usepackage{adjustbox}
\usepackage{siunitx} 
\usepackage{hyperref}
\newcommand{\email}[1]{\href{mailto:#1}{\nolinkurl{#1}}}

\usepackage[normalem]{ulem}

\hypersetup{hidelinks} 
\usepackage[switch]{lineno}

\newcommand{\blue}[1]{{\color{black} #1}}

\def\BibTeX{{\rm B\kern-.05em{\sc i\kern-.025em b}\kern-.08em
    T\kern-.1667em\lower.7ex\hbox{E}\kern-.125emX}}
\begin{document}

\title{
Learning a Kinodynamic Trajectory Manifold \\for Impact-Aware Compliant Catching \\of Fast-Moving Objects

\thanks{This work was supported by the National Natural Science Foundation of China (NSFC) under Grant No. 62403211, in part by Youth S\&T Talent Support Programme of Guangdong Provincial Association for Science and Technology (GDSTA) under Grant No. SKXRC2025092.}
}
\author{\IEEEauthorblockN{1\textsuperscript{st} Guorui Pei}
\IEEEauthorblockA{\textit{College of Robotics} \\
\textit{Taiyuan University of Technology}\\
Taiyuan, China \\
2024000387@link.tyut.edu.cn}
\and
\IEEEauthorblockN{2\textsuperscript{nd} Mengshi Zhang}
\IEEEauthorblockA{\textit{School of Data Science} \\
\textit{City University of Hong Kong (Dongguan)}\\
Dongguan, China \\
72540670@cityu-dg.edu.cn}
\and
\IEEEauthorblockN{3\textsuperscript{rd} Xi Chen}
\IEEEauthorblockA{\textit{School of Advanced Engineering} \\
\textit{Great Bay University}\\
Dongguan, China  \\
chenxi.watermelon@outlook.com}
\and
\IEEEauthorblockN{4\textsuperscript{th} Jinsong Wu}
\IEEEauthorblockA{\textit{Department of Mechanical Engineering} \\
\textit{The Hong Kong Polytechnic University}\\
Hong Kong SAR, China \\
23037993r@connect.polyu.hk}
\and
\IEEEauthorblockN{5\textsuperscript{th} Jiaming Qi}
\IEEEauthorblockA{\textit{College of Mechanical and Electrical Engineering} \\
\textit{Northeast Forestry University}\\
Harbin, China \\
jiamingqi@nefu.edu.cn}
\and
\IEEEauthorblockN{6\textsuperscript{th} Peng Zhou*}
\IEEEauthorblockA{\textit{School of Advanced Engineering} \\
\textit{Great Bay University}\\
Dongguan, China \\
*pzhou@gbu.edu.cn}
}

\maketitle

\begin{abstract}
Fast catching of free-flying objects is difficult because of short reaction time, impact uncertainty, and kinodynamic constraints. We use reinforcement learning in simulation to collect successful catching trajectories and learn a low-dimensional kinodynamic trajectory manifold. At run time, the estimated object initial state is mapped directly to a reference catching trajectory without online nonlinear optimization. The trajectory is tracked with compliant control near contact for improved impact absorption and capture stability.
\end{abstract}


\section{Introduction}
Dynamic catching of fast-moving objects \cite{kim2014catching, salehian2016dynamical,  zhang2025catch} is important for warehouse automation, aerial delivery, service robotics, and robot sports. Unlike static grasping~\cite{lee2026non,zhou2025bagit}, it requires short-horizon reaction, coordinated arm-hand motion, and robustness to impact under varying incoming conditions.

Existing methods \cite{yu2023catch} are typically based on online optimization \cite{zhou2024bimanual, zhou2023neural} or end-to-end policy learning \cite{zhou2026failure, zhou2024imitating}. Optimization can enforce constraints but is often slow in contact-rich interception. End-to-end reinforcement learning \cite{deng2026causality, sutton1998reinforcement} can discover successful behaviors, but direct policy execution \cite{hao2026tla, pan2025multi} offers limited control over whole-trajectory feasibility~\cite{mnih2015human}. A practical solution should combine offline exploration with structured online generation.

\begin{figure*}[htbp]
    \centering
    \includegraphics[width=0.99\textwidth]{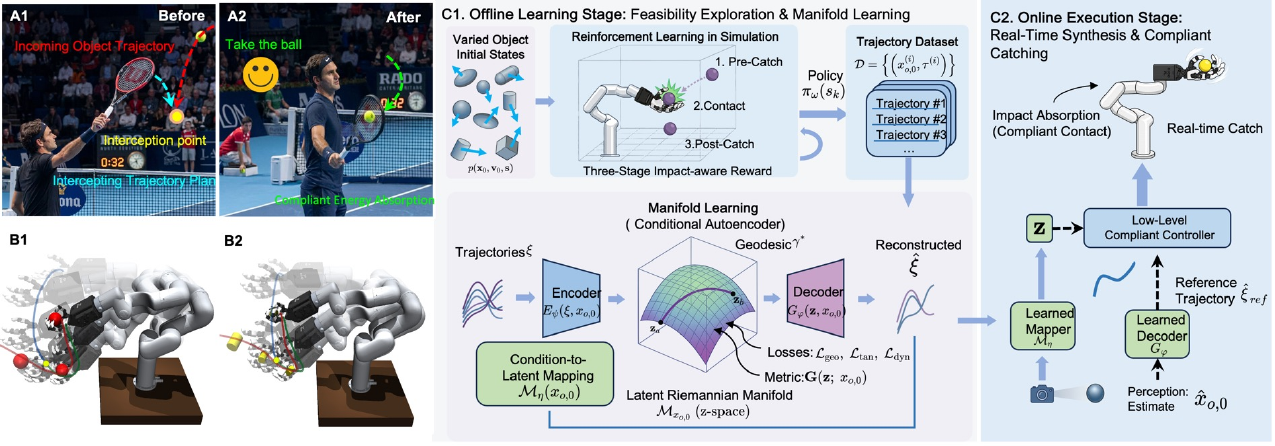}
    \caption{\textbf{Overview of the proposed offline-to-online framework for impact-aware compliant catching.}
    \textbf{A1--A2}: Motivation from human tennis interception—trajectory planning toward the interception point (A1) followed by compliant energy absorption upon contact (A2).
    \textbf{B1--B2}: Simulated robot arm-hand system performing analogous catching motions.
    \textbf{C1 | Offline stage.} RL with a three-stage impact-aware reward explores catching motions over diverse object initial states, producing a trajectory dataset $\mathcal{D}$. A conditional autoencoder then learns a latent Riemannian manifold $\mathcal{M}_{x_{o,0}}$ with geodesic ($\mathcal{L}_{\mathrm{geo}}$, $\mathcal{L}_{\mathrm{tan}}$) and dynamics ($\mathcal{L}_{\mathrm{dyn}}$) regularization, together with a condition-to-latent mapping $M_\eta$.
    \textbf{C2 | Online stage.} Perception estimates the object state $\hat{x}_{o,0}$; $M_\eta$ and the decoder $G_\varphi$ synthesize a reference trajectory $\hat{\xi}_{\mathrm{ref}}$, which a low-level compliant controller tracks for impact absorption and stable capture.}
    \label{fig:framework_overview}
\end{figure*}

Inspired by compliant human interception in tennis (see Fig. \ref{fig:framework_overview}), we propose an offline-to-online framework for impact-aware catching. RL in simulation generates successful arm-hand trajectories over diverse object initial states. A low-dimensional kinodynamic trajectory manifold is then learned and queried online from the estimated object state \cite{paraschos2013probabilistic,arvanitidis2018metric}. The contributions are: (1) an RL pipeline for collecting feasible and impact-aware catching trajectories; (2) a conditioned trajectory manifold for efficient online synthesis; and (3) a trajectory-centric framework combining interception with compliant impact absorption.

\section{Problem Formulation}
We consider the problem of catching a fast-moving object in free flight using a 7-DoF manipulator equipped with a five-finger compliant gripper. Let the object state at time $t$ be
\begin{equation}
x_o(t) = \begin{bmatrix} p_o(t)^\top & v_o(t)^\top \end{bmatrix}^\top \in \mathbb{R}^6,
\end{equation}
where $p_o(t) \in \mathbb{R}^3$ and $v_o(t) \in \mathbb{R}^3$ denote object position and velocity. The robot state is
\begin{equation}
x_r(t) = \begin{bmatrix} q(t)^\top & \dot{q}(t)^\top \end{bmatrix}^\top, \quad u(t) = \tau(t),
\end{equation}
where $q(t) \in \mathbb{R}^7$ and $\tau(t) \in \mathbb{R}^7$ are joint coordinates and torques.

Given an estimated object trajectory $\hat{x}_o(t)$ over $t \in [0,T]$, the goal is to generate a dynamically feasible catching motion reaching a valid interception state at some catching time $t_c \in [0,T]$ while ensuring stable post-impact capture. Let $y_e(q)$ and $\dot{y}_e(q,\dot{q})$ denote the end-effector pose and spatial velocity. The interception condition is
\begin{equation}
y_e(q(t_c)) \approx y_o(t_c), \quad \dot{y}_e(q(t_c),\dot{q}(t_c)) \approx \dot{y}_o(t_c),
\end{equation}
where $y_o(t_c)$ and $\dot{y}_o(t_c)$ are the object pose and velocity in the grasp frame.

To improve robustness under high-speed impact, we optimize a trajectory parameterization $\theta$ on a kinodynamic manifold by minimizing
\begin{equation}
\min_{\theta,\, t_c} J = w_1 J_{\mathrm{track}} + w_2 J_{\mathrm{impact}} + w_3 J_{\mathrm{grasp}} + w_4 J_{\mathrm{effort}},
\end{equation}
subject to
\begin{equation}
\dot{x}_r = f(x_r,u), \quad q \in \mathcal{Q}, \quad \dot{q} \in \dot{\mathcal{Q}}, \quad \tau \in \mathcal{U},
\end{equation}
and geometric feasibility constraints,
\begin{equation}
\mathrm{dist}\big(\mathcal{R}(q(t)),\mathcal{O}(t)\big) \geq 0, \quad t \in [0,t_c),
\end{equation}
where $J_{\mathrm{track}}$, $J_{\mathrm{impact}}$, $J_{\mathrm{grasp}}$, and $J_{\mathrm{effort}}$ penalize interception error, impact severity, poor grasp geometry, and control effort, respectively.

The objective is to learn a low-dimensional trajectory manifold that maps predicted object motion to feasible arm-hand catching motions for real-time impact-aware generation.

\section{Methodology}
The method follows an offline-to-online pipeline. Offline, RL in simulation discovers feasible and impact-aware catching motions over diverse object conditions \cite{schulman2017proximal,haarnoja2018soft}. Online, the learned manifold generates a feasible reference trajectory from the incoming object state for real-time interception and compliant capture.

\subsection{RL-Based Feasible Trajectory Generation}
We first formulate data generation as a finite-horizon Markov decision process. At time step $k$, the policy observes the object state, robot state, and relative interception geometry,
\begin{equation}
s_k = \begin{bmatrix} x_o(k)^\top & x_r(k)^\top & \phi_k^\top \end{bmatrix}^\top,
\end{equation}
where $\phi_k$ includes relative position, relative velocity, and hand-object alignment. The policy outputs an arm-hand control command
\begin{equation}
a_k = \pi_\omega(s_k),
\end{equation}
which may be implemented as target joint increments, joint torques, or another low-level reference.
The reward decomposes into three stages matching the catching phases, plus a penalty:
\begin{equation}
r_k = R_{\mathrm{s1},k} + R_{\mathrm{s2},k} + R_{\mathrm{s3},k} - R_{\mathrm{pen},k}.
\end{equation}

\textbf{Stage~1} (Pre-Catch Tracking) promotes approach and interception alignment before cushioning:
\begin{equation}
R_{\mathrm{s1},k} = \lambda_1 r_{\mathrm{dist}} + \lambda_2 r_{\mathrm{align}} + \lambda_3 r_{\mathrm{progress}},
\end{equation}
where $r_{\mathrm{dist}}$, $r_{\mathrm{align}}$, and $r_{\mathrm{progress}}$ encourage proximity, palm--velocity alignment, and progress toward the interception point.

\textbf{Stage~2} (Soft Contact \& Cushioning) handles force unloading once the object enters the soft zone ($d\!\le\!0.3$\,m) and continues through the post-contact drag buffer:
\begin{equation}
R_{\mathrm{s2},k} = \lambda_4 r_{\mathrm{vel}} + \lambda_5 r_{\mathrm{drag}} + \lambda_6 r_{\mathrm{impact}},
\end{equation}
where $r_{\mathrm{vel}}$, $r_{\mathrm{drag}}$, and $r_{\mathrm{impact}}$ encourage velocity matching and compliant following while penalizing peak contact force.

\textbf{Stage~3} (Post-Catch Stabilization) promotes secure retention after buffering:
\begin{equation}
R_{\mathrm{s3},k} = \lambda_7 r_{\mathrm{grasp}} + \lambda_8 r_{\mathrm{stable}},
\end{equation}
where $r_{\mathrm{grasp}}$ and $r_{\mathrm{stable}}$ favor fingertip enclosure and low residual object--hand motion.
The penalty term enforces safety and smooth impact response:
\begin{equation}
R_{\mathrm{pen},k} =\lambda_{9} p_{\mathrm{impulse}} + \lambda_{10} p_{\mathrm{col}} + \lambda_{11} p_{\mathrm{limit}} + \lambda_{12} p_{\mathrm{drop}},
\end{equation}
where $p_{\mathrm{impulse}}$ penalizes abrupt force variation, and the remaining terms suppress collisions, joint-limit violations, and object drops.

After convergence, we retain successful episodes only and build a dataset
\begin{equation}
\mathcal{D} = \left\{\left(x_{o,0}^{(i)}, \tau^{(i)}\right)\right\}_{i=1}^{N},
\end{equation}
where $x_{o,0}^{(i)}$ is the initial object state and $\tau^{(i)}$ is the corresponding successful catching trajectory. RL therefore serves as an offline feasibility explorer rather than the final online controller.

\begin{figure*}[htbp]
    \vspace{-0.5cm}
    \centering
    \includegraphics[width=\textwidth]{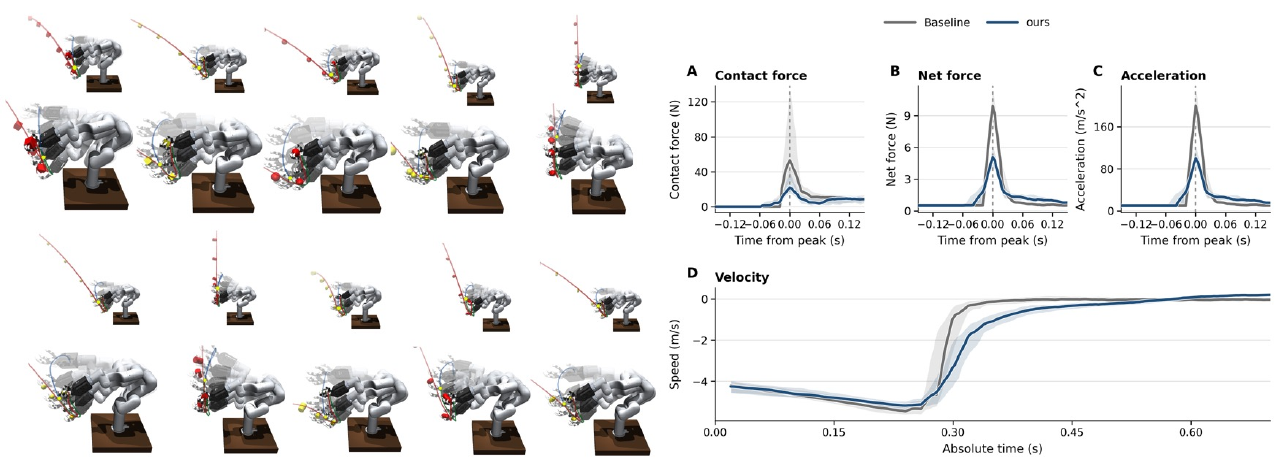}
    \vspace{-0.3cm}
    \caption{
         \textbf{Experimental results.} Left: qualitative catching sequences under diverse incoming trajectories. Right: comparison with the baseline on peak-aligned impact metrics and object vertical velocity, showing smoother compliant capture.
    }
    \label{fig:exp_results}
    \vspace{-0.5cm}
\end{figure*}
\vspace{-0.5cm}

\subsection{Trajectory Representation and Manifold Learning}
RL trajectories generally have different durations and catching times. We therefore align them by the interception instant and resample them to a fixed horizon of $H$ steps. The $i$-th trajectory is represented as
\begin{equation}
\xi^{(i)} = \left[q_1^\top, \dots, q_H^\top, \dot{q}_1^\top, \dots, \dot{q}_H^\top\right]^\top.
\end{equation}
Let $\mathcal{X} \subset \mathbb{R}^{D}$ denote the ambient trajectory space with $D = 14H$. Only a subset satisfies kinodynamic feasibility:
\begin{equation}
\mathcal{F} = \left\{\xi \in \mathcal{X} \,\middle|\, \dot{x}_r = f(x_r,u),\ q \in \mathcal{Q},\ \dot{q} \in \dot{\mathcal{Q}},\ \tau \in \mathcal{U}\right\}.
\end{equation}
Successful catching motions are modeled as a conditional Riemannian manifold $\mathcal{M}_{x_{o,0}} \subset \mathcal{F}$ indexed by object initial state $x_{o,0}$. The manifold is parameterized by a latent variable $z \in \mathbb{R}^{d}$ with $d \ll D$ through
\begin{equation}
\mathcal{M}_{x_{o,0}} = \left\{\xi = G_\varphi(z, x_{o,0}) \mid z \in \mathbb{R}^{d}\right\}.
\end{equation}
This yields a low-dimensional embedded submanifold of the kinodynamic trajectory space.
Given a trajectory $\xi$ and object initial state $x_{o,0}$, the encoder maps the motion into a latent code,
\begin{equation}
z = E_\psi(\xi, x_{o,0}), \quad d \ll \mathrm{dim}(\xi),
\end{equation}
and the decoder reconstructs the trajectory conditioned on the same object state,
\begin{equation}
\hat{\xi} = G_\varphi(z, x_{o,0}).
\end{equation}

The local geometry is characterized by the decoder Jacobian
\begin{equation}
J_G(z, x_{o,0}) = \frac{\partial G_\varphi(z, x_{o,0})}{\partial z} \in \mathbb{R}^{D \times d},
\end{equation}
whose column space defines the tangent space $T_{\xi}\mathcal{M}_{x_{o,0}}$ at $\xi = G_\varphi(z, x_{o,0})$. A pullback Riemannian metric on the latent space is then given by
\begin{equation}
\mathbf{G}(z; x_{o,0}) = J_G(z, x_{o,0})^\top W J_G(z, x_{o,0}),
\end{equation}
where $W \succeq 0$ weights trajectory coordinates. This metric measures latent perturbations by their induced kinodynamic deformation in trajectory space.

Under this metric, the geodesic connecting two latent states $z_a$ and $z_b$ is the minimum-energy curve
\begin{equation}
\gamma^{\ast} = \arg\min_{\gamma(0)=z_a,\,\gamma(1)=z_b} \int_0^1 \dot{\gamma}(s)^\top \mathbf{G}(\gamma(s); x_{o,0}) \, \dot{\gamma}(s) \, ds,
\end{equation}
and the corresponding manifold distance is
\begin{equation}
d_{\mathcal{M}}(z_a,z_b; x_{o,0}) = \left( \int_0^1 \dot{\gamma}^{\ast}(s)^\top \mathbf{G}(\gamma^{\ast}(s); x_{o,0}) \, \dot{\gamma}^{\ast}(s) \, ds \right)^{\frac{1}{2}}.
\end{equation}
This allows interpolation to follow the intrinsic geometry of the feasible set.
The model is trained by minimizing
\begin{equation}
\mathcal{L}_{\mathrm{manifold}} = \mathcal{L}_{\mathrm{rec}} + \lambda_1 \mathcal{L}_{\mathrm{smooth}} + \lambda_2 \mathcal{L}_{\mathrm{dyn}} + \lambda_3 \mathcal{L}_{\mathrm{geo}} + \lambda_4 \mathcal{L}_{\mathrm{tan}},
\end{equation}
where $\mathcal{L}_{\mathrm{rec}} = \|\xi - \hat{\xi}\|_2^2$ penalizes reconstruction error, $\mathcal{L}_{\mathrm{smooth}}$ encourages temporal smoothness, and $\mathcal{L}_{\mathrm{dyn}}$ penalizes dynamic infeasibility. To preserve intrinsic geometry, we introduce a geodesic regularization term
\begin{equation}
\mathcal{L}_{\mathrm{geo}} = \sum_{(i,j) \in \mathcal{N}} \left( d_{\mathcal{M}}\left(z_i,z_j; x_{o,0}^{(i)}\right) - \|\xi^{(i)} - \xi^{(j)}\|_{W} \right)^2,
\end{equation}
which keeps manifold distances consistent with trajectory dissimilarity in the ambient kinodynamic space. We further impose a tangent-space regularization term
\begin{equation}
\mathcal{L}_{\mathrm{tan}} = \sum_{(i,j) \in \mathcal{N}} \left\| \left(\xi^{(j)} - \xi^{(i)}\right) - J_G\left(z_i, x_{o,0}^{(i)}\right)\left(z_j-z_i\right) \right\|_{W}^2,
\end{equation}
where $\mathcal{N}$ denotes local neighbor pairs. Together, $\mathcal{L}_{\mathrm{geo}}$ and $\mathcal{L}_{\mathrm{tan}}$ improve interpolation and online generation.
To enable direct generation from the object initial state, we further learn a condition-to-latent mapping
\begin{equation}
z = M_\eta(x_{o,0}),
\end{equation}
so that the manifold can be queried without running RL online.

\subsection{Online Trajectory Synthesis and Execution}
At test time, perception estimates the incoming object initial state $x_{o,0}$. The latent predictor $M_\eta$ produces a low-dimensional code, and the decoder generates the reference trajectory,
\begin{equation}
\hat{\xi} = G_\varphi\big(M_\eta(x_{o,0}), x_{o,0}\big).
\end{equation}
Equivalently, online generation can be viewed as solving a low-dimensional search problem on the manifold,
\begin{equation}
z^{\ast} = \arg\min_{z \in \mathbb{R}^{d}} \; \ell\big(G_\varphi(z, x_{o,0}), x_{o,0}\big) + \beta C_{\mathrm{dyn}}\big(G_\varphi(z, x_{o,0})\big),
\end{equation}
where $\ell(\cdot)$ measures interception and impact quality, and $C_{\mathrm{dyn}}(\cdot)$ penalizes residual feasibility violations. Because the search is performed in latent space rather than full trajectory space, the dimension is greatly reduced and the solution stays close to the learned feasible manifold.
The latent update can also be written using the Riemannian gradient induced by $\mathbf{G}(z; x_{o,0})$,
\begin{equation}
z_{k+1} = \mathrm{Exp}_{z_k}\left(-\rho \, \mathbf{G}(z_k; x_{o,0})^{-1} \nabla_z \mathcal{J}(z_k)\right),
\end{equation}
where $\mathcal{J}(z) = \ell\big(G_\varphi(z, x_{o,0}), x_{o,0}\big) + \beta C_{\mathrm{dyn}}\big(G_\varphi(z, x_{o,0})\big)$ and $\mathrm{Exp}_{z_k}(\cdot)$ denotes the exponential map on the manifold. This update keeps optimization aligned with the intrinsic geometry of the conditional manifold.

Compared with direct RL execution, this trajectory-centric formulation provides a complete interception motion before contact. Compared with online nonlinear optimization, it shifts heavy computation offline and enables fast trajectory synthesis at run time.

The generated reference is tracked by a low-level compliant controller near contact, improving capture robustness.

\begin{table}[htbp]
    \vspace{-0.3cm}
    \centering
    \caption{Quantitative comparison. Higher success rate is better, while lower force and velocity metrics are preferred.}
    \label{tab:quant_results}
    \scriptsize
    \setlength{\tabcolsep}{2pt}
    \begin{tabular}{>{\centering\arraybackslash}p{1.05cm}>{\centering\arraybackslash}p{1.15cm}>{\centering\arraybackslash}p{1.25cm}>{\centering\arraybackslash}p{1.0cm}>{\centering\arraybackslash}p{1.1cm}}
        \toprule
        Method & \tabincell{c}{Succ. \\ (\%) $\uparrow$} & \tabincell{c}{Peak $F_c$ \\ (N) $\downarrow$} & \tabincell{c}{Net $F$ \\ (N) $\downarrow$} & \tabincell{c}{Peak $v$ \\ (m/s) $\downarrow$} \\
        \midrule
        Baseline & 54.7 & 120.7 & 9.85 & 187.3 \\
        Ours & 85.1 & 32.1 & 4.3 & 83.1 \\
        \bottomrule
    \end{tabular}
    \vspace{-0.3cm}
\end{table}

\section{\blue{Experiments}}
\blue{
\subsection{Experimental Setup}
We evaluate in MuJoCo with a 7-DoF xArm and a five-finger dexterous gripper. As shown in Table~\ref{tab:object_variants} and Fig.~\ref{fig:object_variants}, five geometric primitives at two size scales yield ten object variants (mass $\in[45,55]$\,g). Three domain-randomization pipelines (Table~\ref{tab:domain_rand})---observation, dynamics, and control---are applied to bridge the sim-to-real gap.

\begin{table}[htbp]
    \vspace{-0.2cm}
    \centering
    \caption{\blue{Object geometry variants used in training (5 shapes $\times$ 2 sizes = 10 variants). Mass is uniformly sampled from $[45,\,55]$\,g.}}
    \label{tab:object_variants}
    \scriptsize
    \setlength{\tabcolsep}{3pt}
    \begin{tabular}{lccc}
        \toprule
        \rowcolor{blue!10}
        Shape & Base Size (mm) & Smaller-sized ($\times 0.9$) & Larger-sized ($\times 1.25$) \\
        \midrule
        \rowcolor{blue!10}
        Box        & $25\times25\times25$    & $22.5\times22.5\times22.5$  & $31.3\times31.3\times31.3$ \\
        \rowcolor{blue!10}
        Sphere     & $r{=}27$               & $r{=}24.3$                  & $r{=}33.8$ \\
        \rowcolor{blue!10}
        Ellipsoid  & $30\times20\times25$    & $27\times18\times22.5$      & $37.5\times25\times31.3$ \\
        \rowcolor{blue!10}
        Cylinder   & $r{=}20,\;h{=}25$      & $r{=}18,\;h{=}22.5$        & $r{=}25,\;h{=}31.3$ \\
        \rowcolor{blue!10}
        Capsule    & $r{=}20,\;h{=}20$      & $r{=}18,\;h{=}18$          & $r{=}25,\;h{=}25$ \\
        \bottomrule
    \end{tabular}
    \vspace{-0.2cm}
\end{table}

\begin{figure}[htbp]
    \centering
    \includegraphics[width=\columnwidth]{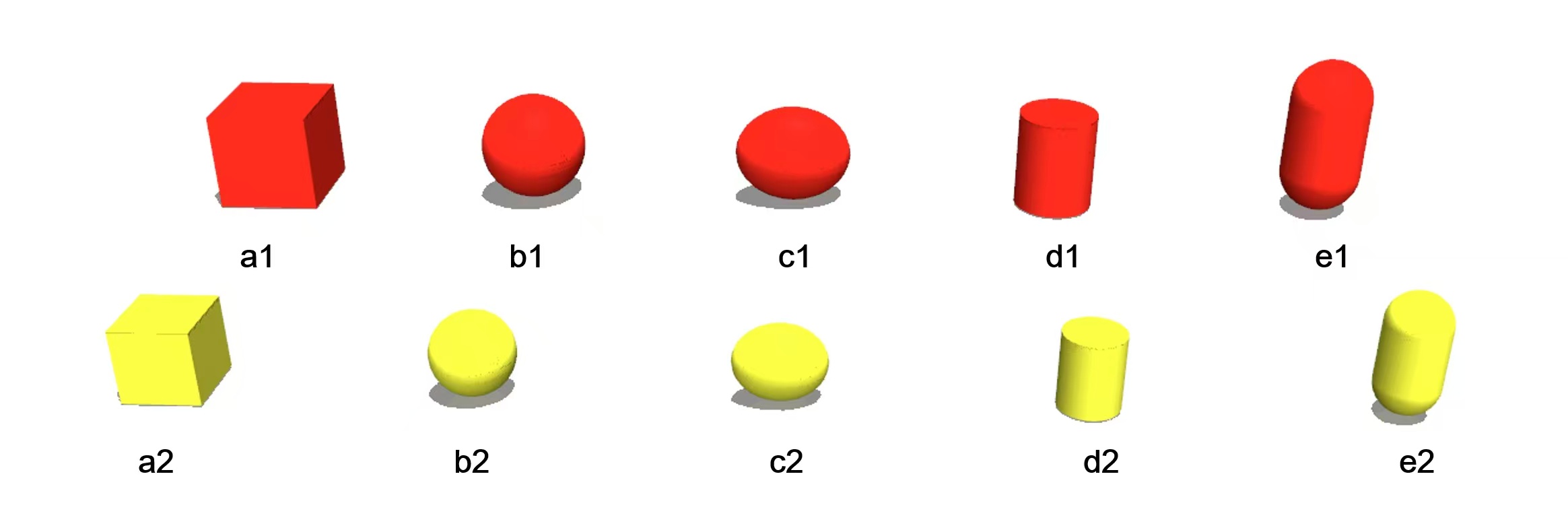}
    \caption{\blue{\textbf{Visualization of the ten object variants used in training.} Top row (red, larger-sized): (a1)~box, (b1)~sphere, (c1)~ellipsoid, (d1)~cylinder, (e1)~capsule. Bottom row (yellow, smaller-sized): (a2)~box, (b2)~sphere, (c2)~ellipsoid, (d2)~cylinder, (e2)~capsule. Dimensions are listed in Table~\ref{tab:object_variants}.}}
    \label{fig:object_variants}
\end{figure}

\begin{table}[htbp]
    \vspace{-0.2cm}
    \centering
    \caption{\blue{Representative training randomization parameters.}}
    \label{tab:domain_rand}
    \scriptsize
    \setlength{\tabcolsep}{3pt}
    \begin{tabular}{llc}
        \toprule
        \rowcolor{blue!10}
        Pipeline & Parameter & Range \\
        \midrule
        \rowcolor{blue!10}
        \multirow{7}{*}{\textbf{Obs.}}
        & Arm joint pos / vel noise      & $0.005$ / $0.01$\,rad \\
        \rowcolor{blue!10}
        & Hand joint pos / vel noise     & $0.015$ / $0.02$\,rad \\
        \rowcolor{blue!10}
        & Object pos / vel noise         & $0.015$\,m / $0.1$\,m/s \\
        \rowcolor{blue!10}
        & Proprio delay                  & $0$--$1$ steps \\
        \rowcolor{blue!10}
        & Vision update period           & $1$--$2$ steps \\
        \rowcolor{blue!10}
        & Vision delay                   & $1$--$2$ steps \\
        \rowcolor{blue!10}
        & Vision dropout prob.           & $0.5$--$1.5$\% \\
        \midrule
        \rowcolor{blue!10}
        \multirow{7}{*}{\textbf{Dyn.}}
        & Joint damping / friction       & $\times(0.9,\;1.1)$ \\
        \rowcolor{blue!10}
        & Joint armature                 & $\times(0.9,\;1.1)$ \\
        \rowcolor{blue!10}
        & Body inertia                   & $\times(0.95,\;1.05)$ \\
        \rowcolor{blue!10}
        & Geom friction                  & $\times(0.95,\;1.05)$ \\
        \rowcolor{blue!10}
        & Actuator gain / bias           & $\times(0.97,\;1.03)$ \\
        \rowcolor{blue!10}
        & Contact solver params          & $\times(0.97,\;1.03)$ \\
        \midrule
        \rowcolor{blue!10}
        \multirow{6}{*}{\textbf{Ctrl.}}
        & Command delay                  & $0$--$1$ steps \\
        \rowcolor{blue!10}
        & Packet loss prob.              & $0$--$0.1$\% \\
        \rowcolor{blue!10}
        & Backlash                       & $0.005$--$0.01$\,rad \\
        \rowcolor{blue!10}
        & Gravity tilt                   & $\pm 0.2^{\circ}$ \\
        \rowcolor{blue!10}
        & Timing jitter                  & $\pm 1$ step \\
        \bottomrule
    \end{tabular}
    \vspace{-0.3cm}
\end{table}

\subsection{Results and Analysis}
Fig.~\ref{fig:exp_results} (left) shows representative catching sequences and the right panel compares peak-aligned impact metrics and object vertical velocity, showing smoother compliant capture. Table~\ref{tab:quant_results} confirms the advantage: our method attains 85.1\% success with peak contact force 32.1\,N versus the baseline's 54.7\% and 120.7\,N.

\subsubsection{Sensitivity to State Estimation Error}
We inject bounded uniform perturbations into the estimated initial object state. As shown in Fig.~\ref{fig:sensitivity}, success remains above 70\% under position perturbations up to 0.05\,m and velocity perturbations up to 0.3\,m/s, owing to the observation-pipeline randomization (Table~\ref{tab:domain_rand}).

\begin{figure*}[htbp]
    \vspace{-0.5cm}
    \centering
    \includegraphics[width=0.95\textwidth]{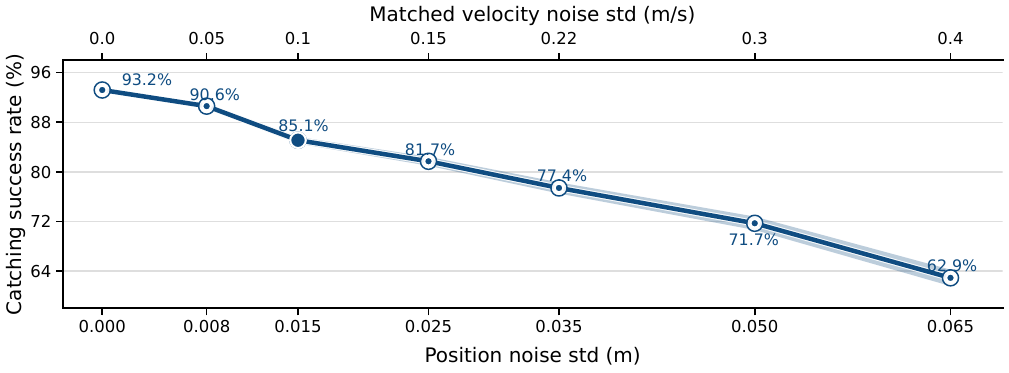}
    \caption{\blue{\textbf{Sensitivity of catching success rate to object initial state estimation error.} Bounded uniform perturbations with a shared perturbation level are jointly applied to the position and velocity components of the estimated initial object state. Shaded regions indicate $\pm 1$ std over five evaluation seeds.}}
    \label{fig:sensitivity}
    \vspace{-0.5cm}
\end{figure*}
}

 \section{\blue{Conclusion}}                                                          
\blue{
This paper presents an offline-to-online framework for impact-aware compliant catching of fast-moving objects. By combining RL-based trajectory exploration with a conditioned kinodynamic manifold, the method enables real-time synthesis of feasible catching motions from the estimated object state without online nonlinear optimization. Experimental results in simulation demonstrate superior catching performance compared to direct policy execution, with higher success rates and significantly reduced impact forces across ten object variants spanning five geometric primitives at two size scales (Fig.~\ref{fig:object_variants}). Sensitivity analysis further confirms that the framework degrades gracefully under realistic levels of state estimation error.

All experiments in this work are conducted in the MuJoCo simulation environment, and real-robot validation is currently lacking. Real-world factors such as actuator dynamics, sensor noise, and imperfect state estimation may introduce challenges not fully captured in simulation, despite the extensive domain randomization employed during training. Future work will focus on real-robot experiments to evaluate the sim-to-real transferability of the learned manifold and the compliant control strategy.
}

\bibliographystyle{IEEEtran}
\bibliography{refs}

\end{document}